\pdfoutput=1

\documentclass[11pt]{article}

\usepackage[]{EACL2023}

\usepackage{times}
\usepackage{latexsym}
\usepackage{graphicx}
\usepackage[T1]{fontenc}
\usepackage[T5]{fontenc}

\usepackage[utf8]{inputenc}

\usepackage{microtype}

\usepackage{inconsolata}

\usepackage{multirow}
\usepackage[normalem]{ulem}
\useunder{\uline}{\ul}{}
\usepackage{threeparttable}

\usepackage{algorithm2e}
\usepackage{xcolor}
\definecolor{navy}{RGB}{30, 8, 142}
\definecolor{darkblue}{rgb}{0, 0, 0.5}

\title{MTet: Multi-domain Translation for English and Vietnamese}

\author{Chinh Ngo$^*$, Trieu H. Trinh$^*$, Long Phan$^*$, Hieu Tran$^*$, 
        \\ {\bf Tai Dang}, {\bf Hieu Nguyen}, {\bf Minh Nguyen} and {\bf Minh-Thang Luong}\\
        VietAI Research}

\begin{document}
\maketitle
\def\thefootnote{*}\footnotetext{The first four authors contributed equally to this work}\def\thefootnote{\arabic{footnote}}
\begin{abstract}
We introduce MTet, the largest publicly available parallel corpus for English-Vietnamese translation. MTet consists of 4.2M high-quality training sentence pairs and a multi-domain test set refined by the Vietnamese research community. Combining with previous works on English-Vietnamese translation, we grow the existing parallel dataset to 6.2M sentence pairs. We also release the first pretrained model EnViT5 for English and Vietnamese languages. Combining both resources, our model significantly outperforms previous state-of-the-art results by up to 2 points in translation BLEU score, while being 1.6 times smaller.
\end{abstract}

\section{Introduction}
Machine Translation is an impactful subdomain of Natural Language Processing that directly benefits the world's interconnected regions and nations, especially so for fast-developing economies such as Vietnam~\cite{vietnamworkingpaper}. Neural machine translation, however, is hindered for many pairs of languages due to their scarce availability. The literature tackling this problem consists mainly of regularization and data augmentation methods~\cite{bpedrop,toan19transformers,kevin18semi}. Recently a more data-centric view with more successful results arises: directly growing the small existing datasets~\citep{m2m100,sat2021ngo,cruz2021improving} and better pretraining methodologies to extract value from large corpora~\citep{mbart,lample2019xlm,song2019mass}.

In this work, we introduce EnViT5, the first pretrained Transformer-based encoder-decoder model for English-Vietnamese, and MTet - \textbf{M}ulti-domain \textbf{T}ranslation for \textbf{E}nglish-Vie\textbf{T}namese, the largest high-quality multi-domain corpus for English-Vietnamese translation of size 4.2M. 
Notably, MTet also focuses on highly technical, impactful yet mostly neglected domains due to their expensive-to-obtain nature such as law and biomedical bitexts. We also introduce a test set of four distinctively different domains, refined and cross-checked by human experts through a data crowdsourcing platform. Our final model, pretrained on EnViT5 and finetuned on MTet + phoMT~\cite{doan-etal-2021-phomt} outperforms previous results by a significant margin of up to 2 points in BLEU score. Finally, we perform experiments to confirm that with the same amount of training data, a multi-domain training set results in a better test performance as shown in Section \ref{mutidomain}, further supporting the multi-domain nature of MTet.



\section{Related Works}
In recent years, research works focusing on improving Machine Translation Systems for Low-Resource Languages have received a lot of attention from both academia and the industry \cite{fbwat19, souretargetmismatch, gu-etal-2018-universal, zulu}. Prior works include collecting more parallel translation data \cite{alt, paracrawl, bicleaner}, training large multilingual models \cite{m2m100, mbart}, and utilizing data augmentation or regularization techniques \cite{selftraining, backtranslateatscale, bpedrop}. Previous works from ParaCrawl \cite{paracrawl} and BiCleaner \cite{bicleaner} focused on mass crawling parallel translation data for many low-resource language pairs. Yet, previous work \cite{phomt} shows that crawling at scale still has limitation and affect downstream translation performance. We also compare our high-quality MTet with other crawling at-scale datasets in Section \ref{data}. 

Encouraging results have also been achieved in low-resource English-Vietnamese translation. The most popular and well-adopted translation dataset for English-Vietnamese is IWSLT15 \cite{cettolo-etal-2015-iwslt}, which consists of 133K text pairs collected from TED talk transcripts. Some studies \cite{provilkov-etal-2020-bpe, 10.5555/3454287.3454681, nguyen-salazar-2019-transformers} show decent improvements through different regularization techniques. Recently, PhoMT \cite{phomt} and VLSP2020 \cite{vlsp2020-mt} released larger parallel datasets of size 3M and 4M text pairs, extracted from publicly available resources for the English-Vietnamese translation. mBART model trained on PhoMT sets the current state-of-the-art results

\begin{table*}[ht]
\centering
\caption{Results on PhoMT English-Vietnamese Translation Test Set}

\begin{threeparttable}

\begin{tabular}{c|c|c|ll|l|l}
\hline
\multirow{2}{*}{Model}       & \multicolumn{1}{l|}{\multirow{2}{*}{\# Params}} & \multirow{2}{*}{Pretrained} & \multicolumn{2}{c|}{Finetuned}                       & \multicolumn{1}{c|}{\multirow{2}{*}{En-Vi}} & \multicolumn{1}{c}{\multirow{2}{*}{Vi-En}} \\ \cline{4-5}
                             & \multicolumn{1}{l|}{}                           &                              & \multicolumn{1}{c|}{Dataset}              & \# pairs & \multicolumn{1}{c|}{}                       & \multicolumn{1}{c}{}                       \\ \hline
M2M100                       & 1.2B                                            & -                            & \multicolumn{1}{l|}{CCMatrix + CCAligned} & 7.5B     & 35.83                                       & 31.15                                      \\ \hline
Google Translate             & -                                               & -                            & \multicolumn{2}{c|}{-}                               & 39.86                                       & 35.76                                      \\ \hline
Bing Translator              & -                                               & -                            & \multicolumn{2}{c|}{-}                               & 40.37                                       & 35.74                                      \\ \hline
Transformer-base             & 65M                                             & -                            & \multicolumn{1}{l|}{PhoMT}                & 3M       & 42.12                                       & 37.19                                      \\ \hline
Transformer-big            & 213M                                            & -                            & \multicolumn{1}{l|}{PhoMT}                & 3M       & 42.94                                       & 37.83                                      \\ \hline
mBART\textsuperscript{$\dagger$}                       & 448M                                            & CC25                         & \multicolumn{1}{l|}{PhoMT}                & 3M       & 43.46                                       & 39.78                                      \\ \hline
\multirow{2}{*}{EnViT5-base} & \multirow{2}{*}{275M}                            
                                                                           & \multirow{2}{*}{CC100}       & \multicolumn{1}{l|}{MTet}                 & 4.2M     & 43.87                                       & {\ul 39.57}                                \\ \cline{4-7} 
                             &                                                 &                              & \multicolumn{1}{l|}{MTet + PhoMT}         & 6.2M     & \textbf{45.47}                              & \textbf{40.57}                             \\ \hline
\end{tabular}
\begin{tablenotes}
      \small
      \item \textit{Notes:} The best scores are in bold and second best scores are underlined. (\textit{$\dagger$}) mBART trained on PhoMT train set are published work \cite{phomt} that previously achieved state-of-the-art results on English-Vietnamese translation. 
\end{tablenotes}
\end{threeparttable}

\label{phomt_results}
\end{table*}

\section{MTet: a Machine Translation dataset in English and Vietnamese}
\label{data}
In this section, we describe in details our MTet - \textbf{M}ultidomain \textbf{T}ranslation for \textbf{E}nglish-vie\textbf{T}namese dataset. We curated a total of 4.2M training examples\footnote{Our work started and progress concurrently to PhoMT, therefore a significant chunk of our data is overlapped. After deduplication, 3M new training examples are contributed on top of PhoMT existing training set.}. 
Based on the curation methodology, we divide this data into four types.

\paragraph{Combining existing sources} This includes sources from the Open Parallel corPUS~\cite{TIEDEMANN12463}, spanning across different domains such as educational videos~\cite{abdelali-etal-2014-amara}, software user interface (GNOME, KDE4, Ubuntu), COVID-related news articles (ELRC), religious texts~\cite{christodouloupoulos2015massively}, subtitles (Tatoeba), Wikipedia~\cite{wolk2014building}, TED Talks~\cite{reimers-2020-multilingual-sentence-bert}. Together with the original IWSLT'15~\cite{iwslt2015} training set, the total dataset reaches 1.2M training examples. 
We train a base Transformer on this data, denoted $bT_A$, to aid the collection of other data sources described below.

\paragraph{Scoring and filtering} Another large source from OPUS is OpenSubtitles~\cite{dataset_opensub} and CCAlign-envi~\cite{elkishky_ccaligned_2020} of sizes 3.5M and 9.3M respectively. For OpenSubtitles, manual inspection showed inaccurate translations similar to the previous observations in \citet{phomt}. Including CCAlign-envi as-is will significantly reduce the model test performance in test set (Appendix \ref{app:ccalign}). For this reason, we make use of $bT_A$ to score each bitext by  computing the loss of all text pairs and select the best 700K training examples using cross-validation on the tst2013 test set\footnote{https://github.com/stefan-it/nmt-en-vi}. CCAlign-envi, on the other hand, is entirely discarded through the same process. 

\paragraph{Dynamic Programming style alignment} Another large source of parallel data but trickier to extract comes from weakly-aligned books and articles~\cite{ladhak2020wikilingua}. This includes many mismatches at sentence and paragraph levels due to versioning, translator formatting, extra headers and page footers information. We propose a dynamic-programming style alignment algorithm detailed in Algorithm~\ref{alg:dp}, a simplified version of BleuAlign~\citep{bleualign}, to filter and align sentences between each pair of documents, maximizing the total BLEU score after alignment. 
In total, we collected 900K training examples from 300 bilingual books and news articles.

\paragraph{Manual crawl and clean} For this source, we focus on more technical and high-impact domains, this include law documents and biomedical scientific articles. We manually crawl and clean across 20 different websites of public biomedical journals and law document libraries, treating them individually due to their significantly different formatting. 
We also manually crawl and clean some other available websites that are more straightforward to process, as detailed in Appendix~\ref{app:source}. Overall, this source contributed another 1.2M training examples. 

\paragraph{Data crowdsourcing for MTet multi-domain test set} We utilize dataset.vn to distribute 4K test examples held out from the collected data to 13 human experts to further refine its content. These domains include biomedical, religion, law, and news. 

Overall, we collected 4.2M training examples across all sources. 
After combining MTet with PhoMT and IWSLT'15, we grew the existing training set from 3M to 6M training examples. Compared to the existing data sources, this dataset is both larger and much more diverse, with the inclusion of technical, impactful, yet so far mostly neglected domains such as law and biomedical data. 

\section{EnViT5}
\subsection{Model}
EnViT5 is a Text-to-Text Transfer Transformer model follows the encoder-decoder architecture proposed by \cite{vaswani2017attention} and the T5 framework proposed by \cite{t5}. The original works of T5 proposed five different configurations in model size: small, base, large, 3B, and 11B. For the practical purpose of the study, we adapt the base architecture for EnViT5 and save the bigger models for future works. 

We train EnViT5 models from scratch with the input and output length of 1024 tokens and batch size of 256. 
For the self-supervised learning objectives, we use the span-corruption objective with a corruption rate of 15\%.

\subsection{Pretraining data}
\label{sec:pretraindata}
We use the CC100 Dataset (Monolingual Datasets from Web Crawl Data) \cite{ccnet} for pre-training the model. The corpus contains monolingual data for over 100 languages. The corpus was constructed using the pipeline provided by \cite{ccnet} through processing January-December 2018 Commoncrawl snapshots.
Following the discussion regarding the importance of long context sequences during pretraining for T5 models from previous works \cite{vit5}, we process and filter for 80GB of long sequence (fit in 1024-length embedding) for each language.



\begin{table*}[]
\centering
\caption{BLEU scores of Transformer\textsubscript{base} on MTet Multi-Domain Test Set}
\begin{tabular}{|c|cccc|cccc|}
\hline
\multirow{2}{*}{Dataset} & \multicolumn{4}{c|}{En-Vi}                                                                                                       & \multicolumn{4}{c|}{Vi-En}                                                                                                      \\ \cline{2-9} 
                         & \multicolumn{1}{c|}{Law}            & \multicolumn{1}{c|}{Religion}       & \multicolumn{1}{c|}{News}           & Medical        & \multicolumn{1}{c|}{Law}            & \multicolumn{1}{c|}{Religion}       & \multicolumn{1}{c|}{News}           & Medical       \\ \hline
300K Ted-talk            & \multicolumn{1}{c|}{16.43}          & \multicolumn{1}{c|}{{\ul 20.55}}    & \multicolumn{1}{c|}{{\ul 27.74}}    & {\ul 14.68}    & \multicolumn{1}{c|}{10.92}          & \multicolumn{1}{c|}{{\ul 18.54}}    & \multicolumn{1}{c|}{{\ul 20.50}}    & 7.61          \\ \hline
300K Law                 & \multicolumn{1}{c|}{{\ul 20.6}}     & \multicolumn{1}{c|}{5.2}            & \multicolumn{1}{c|}{13.07}          & 14.035         & \multicolumn{1}{c|}{{\ul 19.15}}    & \multicolumn{1}{c|}{4.97}           & \multicolumn{1}{c|}{11.275}         & {\ul 12.535}  \\ \hline
Multi-domain             & \multicolumn{1}{c|}{\textbf{22.07}} & \multicolumn{1}{c|}{\textbf{34.77}} & \multicolumn{1}{c|}{\textbf{34.77}} & \textbf{28.76} & \multicolumn{1}{c|}{\textbf{20.45}} & \multicolumn{1}{c|}{\textbf{32.21}} & \multicolumn{1}{c|}{\textbf{28.66}} & \textbf{22.4} \\ \hline
\end{tabular}
\label{table:multidomain}
\end{table*}

\section{Benchmarking EnViT5 and MTet}

\subsection{Experimental settings}

To develop our analysis, we conduct experiments to verify the quality of our MTet dataset and our pretrained bilingual model EnViT5 on both English-to-Vietnamese and Vietnamese-to-English translation. We are interested in the final performance of EnViT5 trained on MTet and PhoMT and aim to demonstrate the best results for both research communities and industry applications.

 We compare EnViT5 against well-known engines and baseline models: Google Translate, Bing Translator, Transformer-base, Transformer-large \cite{vaswani2017attention}, and mBART \cite{phomt}. All our models are trained for 30 epochs with a batch size of 256. We use SacreBLEU \cite{sacrebleu} to compute the case-sensitive BLEU score on 
 the PhoMT test set \cite{phomt}.

\subsection{Results}

Table \ref{phomt_results} presents BLEU scores of our models on both translation directions. A first takeaway is that the large finetuned English-Vietnamese translation dataset accounts for the significant improvement of both En-Vi and Vi-En translations. Both Transformer models \cite{vaswani2017attention} and EnViT5 models \cite{t5} without self-supervised learning steps still achieve notable results on translations compared to current famous translation models from Google Translate and Bing Translator. 


Our EnViT5\textsubscript{base} model when training on a combination of MTet and the released PhoMT achieves state-of-the-art results on low-resource English-Vietnamese translation (\textbf{45.47} and \textbf{40.57} for En-Vi and Vi-En respectively). EnViT5 models outperform current existing multilingual models mBART and M2M100 while being significantly smaller in parameter size (275M parameters compared to 448M and 1.2B). This allows our models not only be able to scale in academia but also very promising for industry and community applications.


\section{Evaluating multi-domain training data}
\label{mutidomain}

In this section, we investigate the importance of multi-domain in training data for a Machine Translation. Since each domain tends to be different in textual structure and style, the ability to generalize across domains will makes translation models more practical in real-world applications.

For fair comparison between different domains, pretraining is not used. 
We start from Transformer\textsubscript{base} \cite{vaswani2017attention} and compare the following three training data on our multi-domain test set described in Section~\ref{data}: (1) 300k Multi-Domain sentence pairs, (2) 300K Ted-talk sentence pairs, and (3) 300K Law sentence pairs.

Besides TED Talk and Law, other domains do not have enough data to fairly take part in our comparison.
The result of this experiment is shown in Table \ref{table:multidomain}. There is a significant increase in BLEU scores across all domains when the model is trained on a Multi-domain training set. Surprisingly, training on Multi-domain data gives better performance on the Law domain than training on the pure Law parallel training dataset itself. 
This result indicates that multi-domain data during supervised training does indeed lead to better test set performance.

\section{A time budget comparison of self-supervised and supervised data}
In this experiment, we first start with IWSLT'15 of 133K training examples and follow two separate processes to improve test performance on top of this initial data point: (1) we pretrain the model on an amount of non-aligned bilingual texts described in section~\ref{sec:pretraindata} before further fine-tuning it on the IWSLT'15 training set {\it for one epoch}; (2) we simply grow the IWSLT'15 training set by an amount of high-quality parallel text before training {\it for one epoch} from random weights.

\begin{figure}[ht!]
    \centerline{\includegraphics[width=0.5\textwidth]{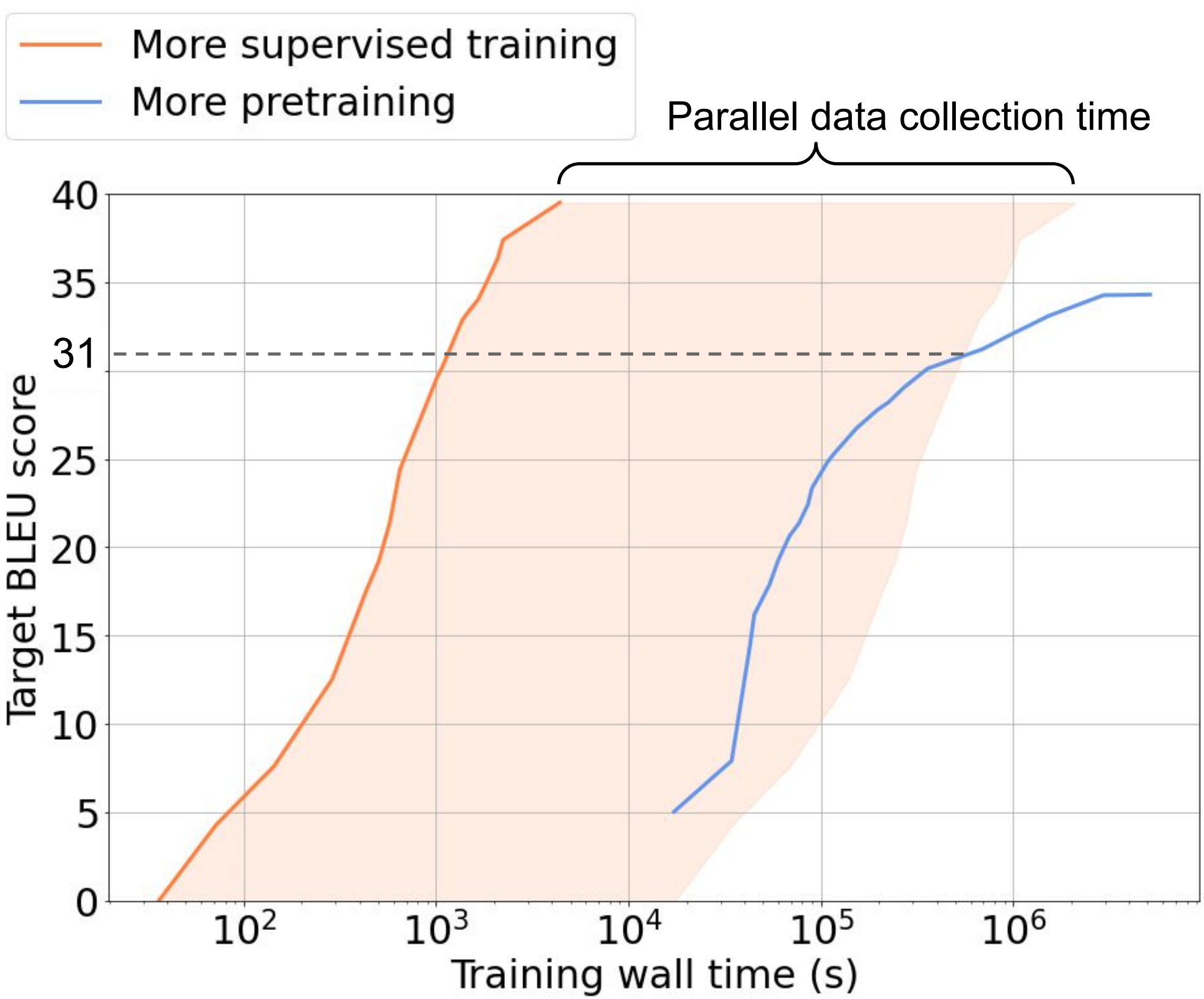}}
        \caption{Improvement on 133k bitexts}
    \label{fig:coord}
\end{figure}

In both methods, we measure the improvement in BLEU score at various amounts of additional data. Following this, we are able to measure the amount of training wall time needed to achieve the target BLEU score. This time is also directly proportional to the added amount of data. 

As reported in Figure ~\ref{fig:coord}, we first confirmed that BLEU score on the test set steadily improved as both types of data grows, albeit at vastly different rates. 
BLEU scores improvement from pretraining quickly diminishes, eventually hitting a wall. After this point, it becomes infeasible to reach further target BLEU scores by pure pretraining, a 1.5X increase in pretraining data does not lead to any meaningful improvement. At a target BLEU score of 34, we found that it took close to 1000X the amount of data and 2000X training wall time for pretraining to reach the same performance as supervised training.

\section{Conclusion}
In this work, we released a state-of-the-art pretrained Transformer model and the largest multi-domain parallel dataset for English-Vietnamese translation. Namely,
MTet consists of 4.2M high-quality training sentence pairs collected using various methods across multiple domains of data. Combining with phoMT, the total training data grow to 6.2M sentence pairs, currently the largest publicly available dataset. Further, we released EnviT5, the first pretrained model for English and Vietnamese languages. Finetuning EnviT5 on MTet, we obtained state-of-the-art results with improvements up to 2 points in BLEU score for English-Vietnamese Translation and 1 BLEU score in Vietnamese-English translation. With much better test results, our model is also 1.6 times smaller than previous translation models with much faster inference time.



\section{Limitations}
Although we conjecture that behaviors observed in our work will exhibit similarly in other low-resource language pairs, there are legitimate reasons to believe different languages might behave differently due to their own unique morphology. Generalizing our work to other pairs requires nontrivial effort and we leave this for future investigation.


\section{Acknowledgements}
We would like to thank the Google TPU Research Cloud (TRC) program, Soonson Kwon (Google ML Ecosystem programs Lead) and Ba Ngoc Nguyen (Google Developer Experts in ML) for their support. This project also receives generous support from Cohost.ai, specifically Mr. Kim Cuong Pham, and dataset.vn, for managing and labeling our multi-domain test data. We appreciate the effort of volunteers who refine the multi-domain test dataset: Hàn Thọ Hoà, Trần Viết Đình, Dương Ngọc Doanh, Nguyễn Bùi Thiên Anh, Nguyễn Công Khanh, Triệu Khắc Đức, Trần Thị Lành, Hoàng An, Hữu Doanh, Ngọc Khánh, Trọng Văn, Huỳnh Anh, Thu Huyền.

\bibliography{anthology,custom}
\bibliographystyle{acl_natbib}

\appendix

\section{Dataset Statistics}

\begin{figure*}[ht!]
    \centering
    \includegraphics[width=\textwidth]{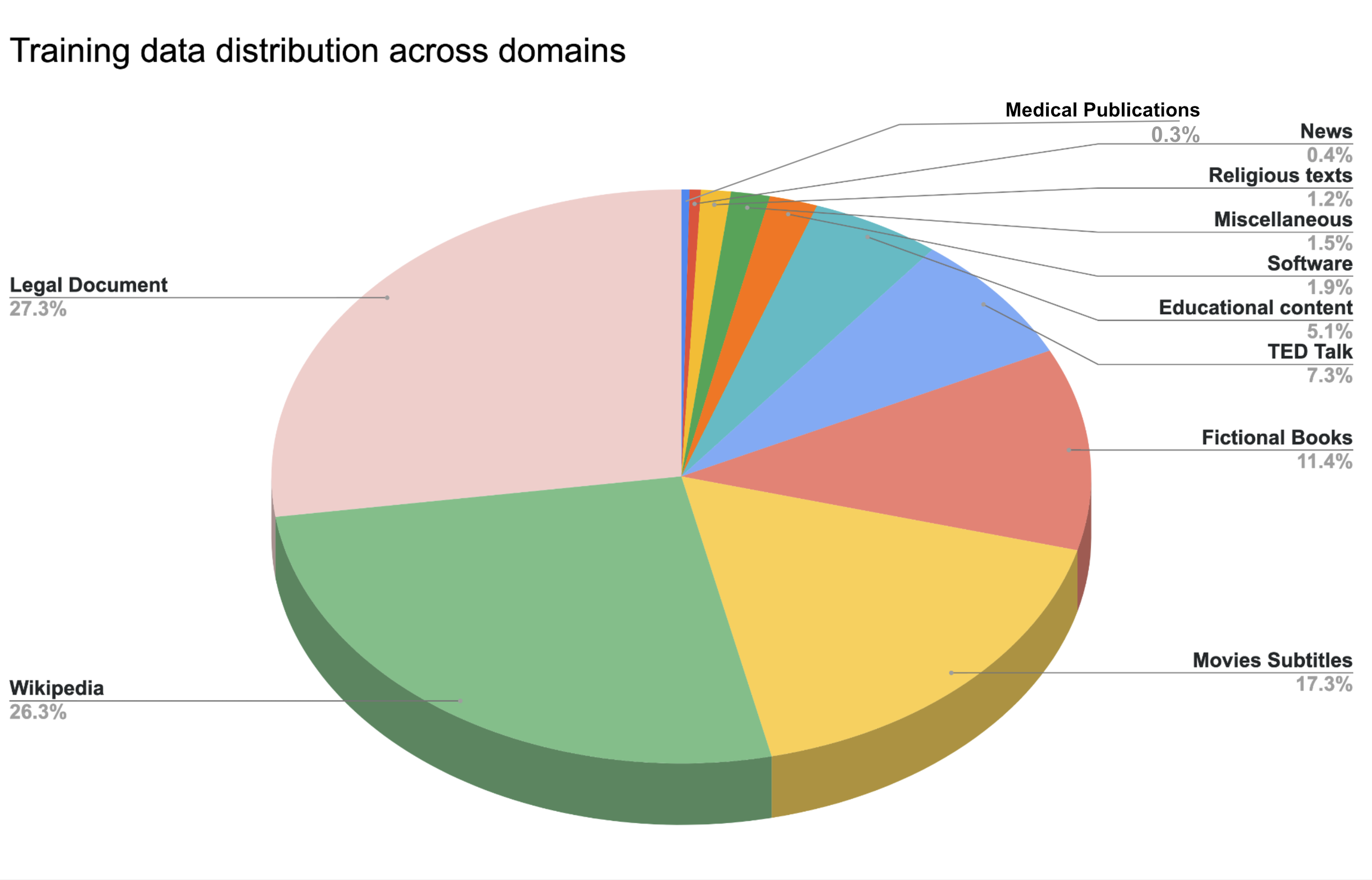}

    \caption{Training data distribution across multiple domains}
    \label{fig:data_distribution}
\end{figure*}

The data distribution of our MTet dataset is described in Figure 
\ref{fig:data_distribution}.

\begin{algorithm}[tb!]
\small
\caption{Alignment algorithm for weakly-aligned pairs of documents. The algorithm strips away a portion of sentences in each document and matches the remaining sentences into pairs, aiming to maximize the total BLEU score with respect to a given translation model.}\label{alg:three}
\label{alg:dp}
\hrulefill\\
\KwData{\\
($l_e$, $l_v$) : a weakly-aligned pair of documents.\\
$l_e =$ ordered list of $N$ English sentences.\\
$l_v =$ ordered list of $M$ Vietnamese sentences.\\
$t_{src\rightarrow dst}:$ translation model from $src$ to $dst$.}
\KwResult{\\
$p = $ ordered list of aligned text pairs $(e\in l_e, v \in l_v)$ \\
that maximizes $\sum_{(e, v) \in p} \texttt{s}(e, v)$, where \\
$\texttt{s}(e, v) = \texttt{BLEU}(e, t_{vi\rightarrow en}(v)) + \texttt{BLEU}(t_{en\rightarrow vi}(e), v)$
}
\hrulefill \\
 Initialize table \texttt{dp}[0 .. $M$, 0 .. $N$] with $0$s\;
 \For {$m = 1 \rightarrow M$} {
    \For {$n = 1 \rightarrow N$} {
    \texttt{dp}[$m$, $n$] = \texttt{max}(  \\
    \ \ \ \ \texttt{dp}[$m-1$, $n$]  \\
    \ \ \ \ \texttt{dp}[$m$, $n-1$] \\
    \ \ \ \ \texttt{dp}[$m-1$, $n-1$] + \texttt{s}($l_e[m]$, $l_v[n]$) \\ );
}}
 $m = M$\;
 $n = N$\;
 $p = []$\;
 \While{$m > 1, n > 1$} {
    \uIf{case 1}{$m$ = $m-1$}
    \uElseIf{case 2}{$n$ = $n-1$}
    \uElse{add pair ($l_e[m], l_v[n]$) to $p$\; $m$ = $m-1$\; $n$ = $n-1$\;
    }
 }
 \Return{p}\;
 \hrulefill
\end{algorithm}

\section{Data collection time}

\begin{figure*}[ht!]
    \centering
    \includegraphics[width=\textwidth,keepaspectratio]{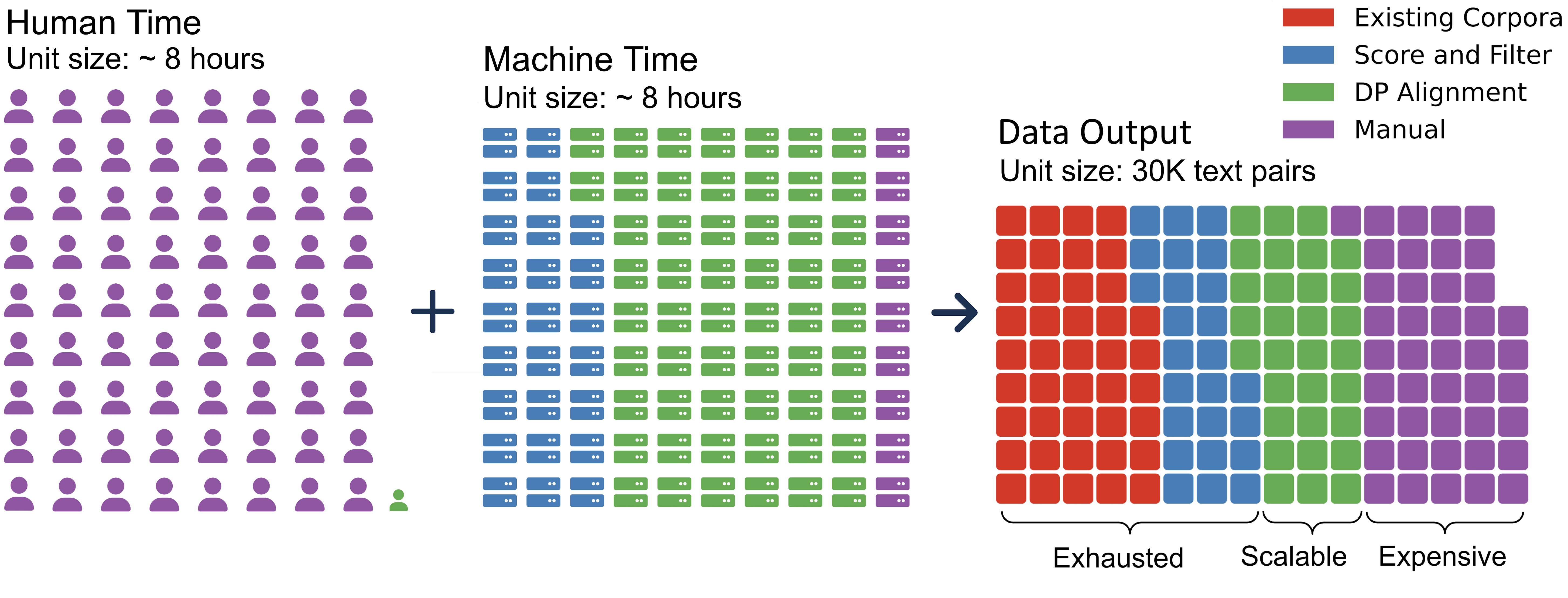}
    \caption{Time required to 4.2M bitexts, color-coded for four tiers of data sources (1) combine existing open-sourced corpora, (2) score and filter noisy sources, (3) DP alignment from weakly-aligned documents, and (4) manual crawl and clean. With comparable outputs, the time invested is vastly different between them. The most expensive approach is manual crawl and clean, while the most scalable is DP alignment.}
    \label{fig:datatime}
\end{figure*}

We record human time as the time spent developing different code bases for crawlers, inspecting manually, cleaning of different data sources, aggregating website sources, and converting files to appropriate text format. Machine time is execution time for long-running jobs such as crawling and rendering millions of websites, batch downloading files, preprocessing large volumes of texts, running inference for millions of sentences on Transformer models, and computing BLEU scores between billions of pairs of sentences. The recorded time is shown in Figure~\ref{fig:datatime}

\section{Quality of existing BiText Mining Datasets}
\label{app:ccalign}
MultiCCAligned \cite{elkishky_ccaligned_2020} massively crawled the Web and aligned bilingual texts using the auto-metric of embedding-based document similarity. This results in 9.3M English-Vietnamese text pairs - the largest collection available to the public at the moment\footnote{The MultiCCAligned paper reported 12.4M pairs, we detected and removed duplicates, which accounted for nearly one quarter of their released data.}. However, auto-metric-based alignment produces data of lower quality than our carefully hand-curated collection, many pairs in MultiCCAligned are themselves low-quality machine translated. Training on MultiCCAligned, therefore, gives a much lower BLEU score, while incorporating MultiCCAligned into our own data slightly decreases our result.

\begin{figure}[ht!]
    \centering
    \includegraphics[width=0.5\textwidth,keepaspectratio]{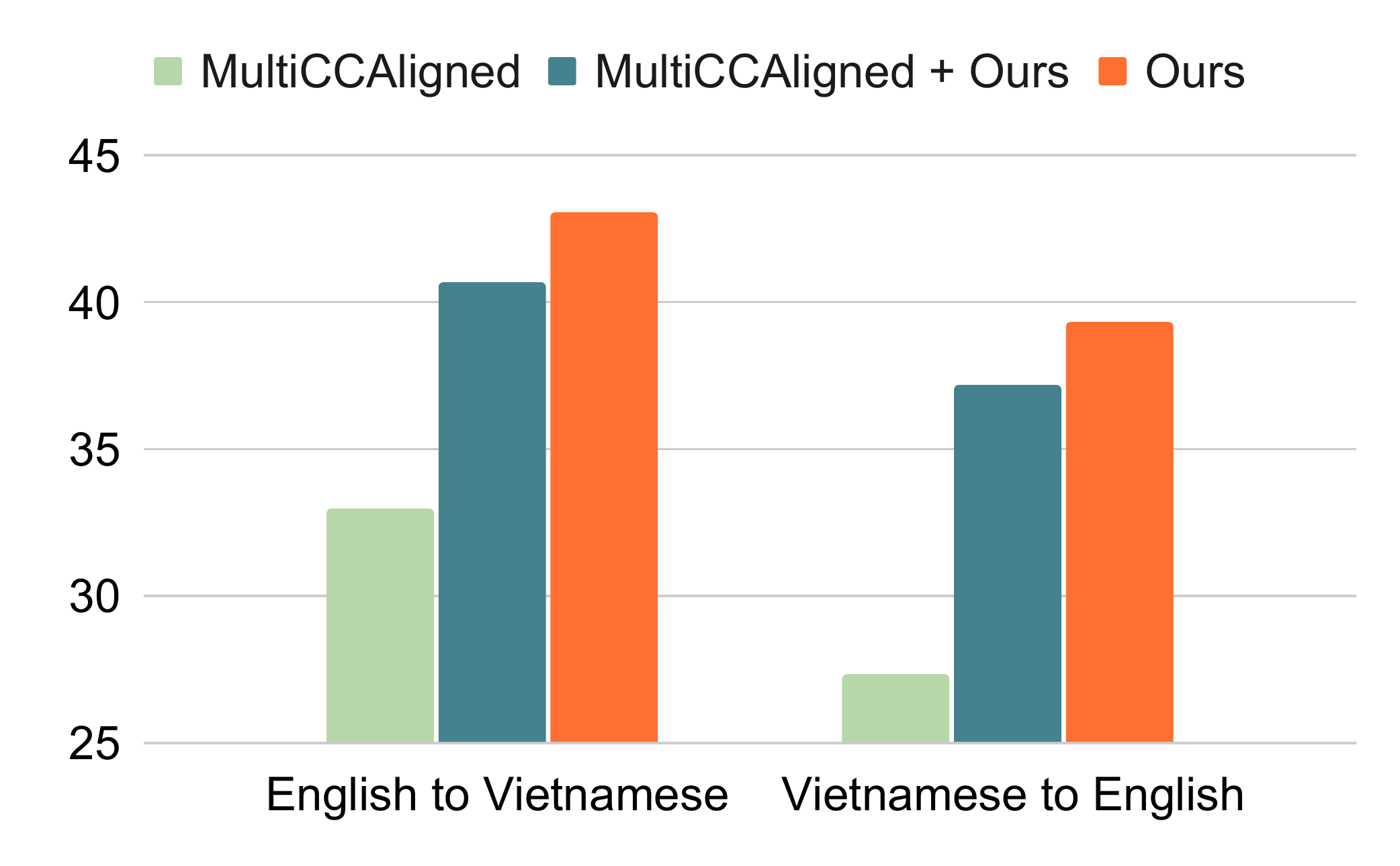}
    \caption{Performance comparison between parallel datasets}
    \label{fig:bitext_quality}
\end{figure}

\section{Data sources for Manual Crawl and Clean}
\label{app:source}


Medical
\begin{itemize}
\color{darkblue}
\item \textsl{https://yhoctphcm.ump.edu.vn}
\item \textsl{http://jmp.huemed-univ.edu.vn}
\item \textsl{http://tonghoiyhoc.vn}
\item \textsl{http://hoinhikhoavn.com}
\item \textsl{http://hoiyhoctphcm.org.vn}
\item \textsl{https://jns.vn}
\item \textsl{https://jprp.vn}
\item \textsl{http://hocvienquany.edu.vn}
\item \textsl{https://sinhlyhoc.com.vn}
\item \textsl{https://tapchinghiencuuyhoc.vn}
\item \textsl{http://tapchi.vienbongquocgia.vn}
\item \textsl{http://vienduoclieu.org.vn}
\item \textsl{https://vjpm.vn/index.php}
\item \textsl{http://vjfc.nifc.gov.vn}
\item \textsl{https://vjs.ac.vn}
\item \textsl{http://vutm.edu.vn}
\item \textsl{https://jcmhch.com}
\item \textsl{http://www.yhth.vn}
\item \textsl{https://sj.ctu.edu.vn}
\item \textsl{https://radiology.com.vn}
\item \textsl{https://vjol.info.vn}
\item \textsl{http://www.vjph.vn}
  
\end{itemize}
Others websites
\begin{itemize}
\color{darkblue}
\item \textsl{https://vietanhsongngu.com }
\item \textsl{https://baosongngu.com}
\item \textsl{https://sachsongngu.top}
\item \textsl{https://tvpl.vn}
\item \textsl{http://vbpl.vn}
\item \textsl{http://automation.net}
\item \textsl{http://tapchixaydungbxd.vn}
\item \textsl{https://duytan.edu.vn}
\item \textsl{https://tapchikhcn.haui.edu.vn}
\item \textsl{https://tapchivatuyentap.tlu.edu.vn}
\item \textsl{http://tapchimoitruong.vn}
\item \textsl{https://translations.launchpad.net}
\item \textsl{https://translationproject.org}
\item \textsl{https://issuu.com/}
\item \textsl{https://lyricstranslate.com}
\item \textsl{https://www.wikihow.com}
\item \textsl{https://d2l.aivivn.com}
\end{itemize}
Youtube Channels
\begin{itemize}
\color{darkblue}
\item \textsl{Khan Academy }
\item \textsl{Ted Ed}
\item \textsl{Asap Science}
\item \textsl{Crash courses}
\item \textsl{GCP Grey}
\item \textsl{Veritasium}
\item \textsl{Vsauce}  
\end{itemize}

\end{document}